\documentclass[sigconf]{acmart}

\copyrightyear{2025}
\acmYear{2025}
\setcopyright{cc}
\setcctype{by}
\acmConference[GeoGenAgent '25]{The 1st ACM SIGSPATIAL International Workshop on Generative and Agentic AI for Multi-Modality Space-Time Intelligence}{November 3--6, 2025}{Minneapolis, MN, USA}
\acmBooktitle{The 1st ACM SIGSPATIAL International Workshop on Generative and Agentic AI for Multi-Modality Space-Time Intelligence (GeoGenAgent '25), November 3--6, 2025, Minneapolis, MN, USA}
\acmDOI{10.1145/3764915.3770724}
\acmISBN{/2025/11}

\AtBeginDocument{%
  }

\usepackage{pifont}
\newcommand{\cmark}{\ding{51}}
\newcommand{\xmark}{\ding{55}}

\begin{document}

\title{From Queries to Insights: Agentic LLM Pipelines for Spatio-Temporal Text-to-SQL}

\author{Manu Redd}
\email{manuredd@ku.edu}
\affiliation{%
  \institution{University of Kansas}
  \city{Lawrence}
  \country{United States}}

\author{Tao Zhe}
\email{taozhe@ku.edu}
\affiliation{%
  \institution{University of Kansas}
  \city{Lawrence}
  \country{United States}}

\author{Dongjie Wang}
\authornote{Corresponding author}
\email{wangdongjie@ku.edu}
\affiliation{%
  \institution{University of Kansas}
  \city{Lawrence}
  \country{United States}}

\begin{abstract}
Natural-language-to-SQL (NL-to-SQL) systems hold promise for democratizing access to structured data, allowing users to query databases without learning SQL. Yet existing systems struggle with realistic spatio-temporal queries, where success requires aligning vague user phrasing with schema-specific categories, handling temporal reasoning, and choosing appropriate outputs. We present an \emph{agentic pipeline} that extends a naive text-to-SQL baseline (defog/llama-3-sqlcoder-8b) with orchestration by a Mistral-based ReAct agent. The agent can plan, decompose, and adapt queries through schema inspection, SQL generation, execution, and visualization tools. 

We evaluate on 35 natural-language queries over the NYC and Tokyo check-in dataset, covering spatial, temporal, and multi-dataset reasoning. The agent achieves substantially higher accuracy than the naive baseline (\textbf{91.4\%} vs.\ \textbf{28.6\%}) and enhances usability through maps, plots, and structured natural-language summaries. Crucially, our design enables more natural human–database interaction, supporting users who lack SQL expertise, detailed schema knowledge, or prompting skill. We conclude that agentic orchestration, rather than stronger SQL generators alone, is a promising foundation for interactive geospatial assistants.
\end{abstract}



\maketitle

\section{Introduction}

Natural-language interfaces (NLIs) to databases promise to make structured data accessible to a far broader range of users by allowing questions to be posed in everyday language rather than Structured Query Language (SQL). This paradigm is especially compelling in domains involving spatio-temporal data, such as location-based social networks (LBSNs), where analysts may wish to investigate mobility trends, venue popularity, or neighborhood-level activity patterns. Despite rapid progress in large language model (LLM)-driven text-to-SQL (NL-to-SQL) systems, current approaches remain fragile when faced with the complexities of real-world geospatial datasets.

Three recurring challenges illustrate the limitations of existing systems.  
\textbf{First, semantic mismatch.} User queries are rarely aligned with database schema terms: people ask for "laundromats," while the schema may contain "Laundry Service"; they inquire about "nightlife," while the database encodes categories such as "Bar", "Nightclub", or "Music Venue". Without mechanisms for adaptation, naive NL-to-SQL systems fail to reconcile intent with schema.  
\textbf{Second, temporal reasoning.} LBSN queries often involve nontrivial temporal structure—hour-of-day trends, weekday/weekend splits, or seasonal windows. These require careful handling of edge cases, such as ranges that cross midnight or holiday-specific comparisons, which most NL-to-SQL systems are not designed to support.  
\textbf{Third, spatial semantics.} Analysts frequently reference neighborhoods, boroughs, or landmarks, none of which are directly encoded in database columns. Accurate analysis requires either injecting external knowledge (e.g., geographic bounding boxes, landmark coordinates) or decomposing queries into subproblems that align with available schema.

Beyond correctness, \textbf{usability remains a fundamental gap}. Naive text-to-SQL systems generally return raw tabular results, leaving interpretation to the user. For non-experts unfamiliar with the schema, SQL, or visualization libraries, this limits practical value: an analyst may know \emph{what} they want to ask but lack the expertise to determine \emph{which table to query}, how to phrase filters, or how best to interpret results. An effective system must therefore mediate between underspecified natural-language intent and concrete database operations, while delivering outputs in forms that aid sensemaking—plots, maps, and concise textual summaries.

\textbf{Positioning.} Our work situates itself at the intersection of text-to-SQL with LLMs (e.g., SQLCoder~\cite{defog2024sqlcoder}), tool-augmented reasoning frameworks (ReAct~\cite{yao2023react}, LangChain~\cite{chase2022langchain}, LangGraph~\cite{langgraph2024}), and geospatial analytics on LBSN datasets (e.g., NYC and Tokyo check-ins~\cite{yang2015modeling}). While each line of work has matured independently, prior research rarely evaluates an \emph{agentic} pipeline tailored for spatio-temporal analysis—one that integrates schema discovery, multi-step planning, external knowledge injection, and task-appropriate visualization into an interactive loop.

\textbf{Our approach.} We present an agentic NL-to-SQL pipeline that embeds a naive SQL generator as a tool inside a Mistral-based ReAct agent. This agent operates in a plan–act–observe loop, dynamically retrieving schema, generating and refining SQL, executing queries, and producing visualizations and summaries. In doing so, it extends the naive pipeline—which simply maps a query to a single SQL statement—by providing error recovery, multi-step decomposition, and interpretability. The result is a system that not only achieves higher accuracy but also supports more natural human–database interaction for users who lack SQL expertise or detailed schema knowledge.

\textbf{Contributions.} This paper makes three contributions:
\begin{enumerate}
    \item We design an agentic NL-to-SQL pipeline for spatio-temporal queries over the NYC and Tokyo check-in dataset, extending a naive baseline with schema grounding, multi-step reasoning, and visualization support.
    \item We provide a head-to-head evaluation against the naive baseline, showing large gains in accuracy (\textbf{91.4\%} vs.\ \textbf{28.6\%}) despite both relying on the same underlying SQL generator.
    \item We analyze failure modes, usability benefits, and trade-offs, distilling practical guidelines for building agentic NL-to-SQL systems that improve both correctness and human–database interaction.
\end{enumerate}

\section{Related Work}

\paragraph{Text-to-SQL.}
Recent progress in natural-language interfaces to databases has been driven by large language models and specialized systems such as SQLCoder~\cite{defog2024sqlcoder}. These approaches achieve strong results on benchmarks like Spider~\cite{yu2018spider} but often fail on spatio-temporal queries involving compositional reasoning, schema mismatch, or vague domain-specific phrasing. Outputs are typically returned as raw tables, leaving interpretation and visualization to the user. \textbf{Gap:} Few works study how to transform underspecified geospatial questions into multi-step analyses with schema grounding, error recovery, and user-ready outputs.

\paragraph{Agentic AI.}
Tool-augmented frameworks such as ReAct~\cite{yao2023react}, LangChain~\cite{chase2022langchain}, and LangGraph~\cite{langgraph2024} allow LLMs to plan, call tools, and adaptively refine their reasoning. These have shown benefits in tasks requiring decomposition, retrieval, and verification. However, evaluations typically focus on generic QA or knowledge tasks. \textbf{Gap:} There is limited evidence on how agentic orchestration improves correctness and usability in NL-to-SQL for spatio-temporal data, or on the marginal impact of specific tools such as schema inspection, SQL execution, or visualization.

\paragraph{Geospatial AI and LBSNs.}
Studies of location-based social network (LBSN) data (e.g., NYC check-ins~\cite{yang2015modeling}) have revealed spatial-temporal regularities in human mobility and activity. Most of this work assumes expert-driven pipelines with handcrafted queries and visualizations. Interactive assistants that accept free-form natural language, resolve schema mismatches, and return geovisualizations remain underexplored. \textbf{Gap:} Prior studies have not systematically evaluated agentic NL-to-SQL pipelines on spatio-temporal LBSN data, especially with per-question correctness and user-facing usability analysis.

\noindent\textbf{Our contribution} is to fill these gaps by (i) instantiating a geospatially-aware ReAct agent that embeds a naive NL-to-SQL model as a tool, and (ii) evaluating it head-to-head against the baseline on 35 spatio-temporal queries, reporting both accuracy and usability improvements.

\section{Dataset}
We evaluate on the public \textbf{NYC and Tokyo LBSN Check-in Dataset} of Yang et~al.~\cite{yang2015modeling}, which was originally introduced to study the spatial-temporal regularity of user activity in location-based social networks (LBSNs). This dataset contains check-ins collected over roughly ten months (2012-04-12 to 2013-02-16), with each check-in annotated by a timestamp, GPS coordinates, and a fine-grained venue category. In total, it consists of:
\begin{itemize}
    \item \textbf{227,428 check-ins in New York City}
    \item \textbf{573,703 check-ins in Tokyo}
\end{itemize}

\noindent Each record captures a user identifier, place identifier, latitude, longitude, semantic category name, and timestamp. These features make the dataset well-suited for spatio-temporal analysis at the intersection of geography (where check-ins occur), time (when activity peaks occur), and semantics (which venue categories dominate behavior).

\subsection{Experimental Setup}
For the purposes of this paper, we deploy the dataset into a relational structure hosted on Supabase with two PostgreSQL tables:
\begin{itemize}
    \item \texttt{checkins\_nyc}: all New York City check-ins
    \item \texttt{checkins\_tokyo}: all Tokyo check-ins
\end{itemize}

\noindent Both tables share an identical schema:
\begin{verbatim}
user_id TEXT,
place_id TEXT,
latitude REAL,
longitude REAL,
category_name TEXT,
checkin_time TIMESTAMP
\end{verbatim}

\noindent We focus on these six fields since they directly support the types of natural-language-to-SQL queries studied in this paper. Other fields available in the dataset (e.g., UTC offsets, normalized timestamps) were omitted for brevity, as they are not essential to our evaluation. This setup enables both \emph{single-city} queries (within the \texttt{checkins\_nyc} table) and \emph{cross-city comparisons} (requiring synthesis across the \texttt{checkins\_nyc} and \texttt{checkins\_tokyo} tables).  

\noindent This dual-table design is particularly important for the evaluation of \textbf{multi-dataset synthesis queries} (e.g., “Compare the most popular categories in NYC vs Tokyo”), which test whether an agent can dynamically reason across multiple databases. These cases (see Q31--Q35 in Table~\ref{tab:per-question}) represent some of the most challenging query types in our benchmark.

\begin{figure*}[t]
  \centering
  \includegraphics[width=\textwidth]{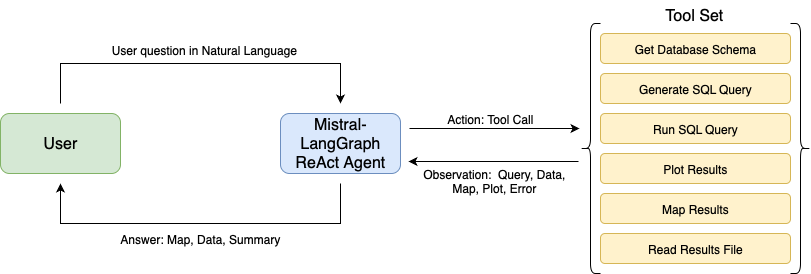}
  \caption{Agentic pipeline for spatio-temporal Text-to-SQL. The ReAct-style agent plans, calls tools (schema, SQL generation, execution, file read, plotting, mapping), observes results/errors, and returns a summarized answer with links to figures/maps. The "Action" and "Observation" arrows in the diagram between the agent and the tool set represent the ReAct chain of thought loop where the agent plans, acts, and observes repeatedly. The Generate SQL Query tool in the tool set is the same llama-3-sqlcoder-8b model that makes the backbone of the naive pipeline.}
  \Description{A flow diagram with a user on the left, an agent in the center using six tools, a database and filesystem for results, and arrows showing a plan–act–observe loop and final answer output.}
  \label{fig:agentic-pipeline}
\end{figure*}

\section{System Design}

Our system architecture consists of two pipelines: a \textbf{naive NL-to-SQL baseline} and an \textbf{agentic pipeline}. Importantly, the agentic pipeline is not a replacement but an \emph{extension} of the naive one: the naive LLM-to-SQL model serves as a callable tool inside the agent’s reasoning loop. This design isolates the marginal value of orchestration, planning, adaptation, and tool use.

\subsection{Naive Pipeline}
We adopt \texttt{defog/llama-3-sqlcoder-8b}~\cite{defog2024sqlcoder} as a representative single-pass text-to-SQL baseline. The pipeline is straightforward:
\begin{enumerate}
    \item Take the user’s natural-language question.
    \item Prompt SQLCoder with the question and schema description.
    \item Return a single SQL query.
    \item Execute the SQL and display the raw tabular result.
\end{enumerate}

This single-pass architecture simplifies deployment and minimizes per-query latency, but it suffers from several limitations:
\begin{itemize}
    \item \textbf{Semantic mismatch.} The baseline model directly translates text to SQL without awareness of dataset-specific vocabulary. As a result, it fails when user terms do not exactly align with schema values. This prevents it from handling synonyms, aliases, or domain-specific phrasing.

    \item \textbf{Robustness limitations.} The naive pipeline is fragile in execution. It often \emph{hallucinates} non-existent columns or functions, produces SQL that does not parse, or mishandles temporal edge cases (e.g., time ranges crossing midnight). Importantly, it cannot adapt to dataset idiosyncrasies (such as inconsistent category naming) or incorporate simple external priors (e.g., holiday dates, geographic boundaries). When queries fail or return empty results, the system has no built-in mechanism for error detection, retry, or refinement.

    \item \textbf{Usability limitations.} Output is limited to raw tables. While static visualization could theoretically be bolted on (e.g., plotting the result of a fixed query), the pipeline cannot \emph{reason dynamically} about when visualization is appropriate, what type (map vs.\ bar chart vs.\ line plot) best fits the task, or how to summarize results in natural language. Consequently, the system provides little support for interpretability or user-centered exploration beyond the literal query execution.
\end{itemize}

Because of these limitations, the naive pipeline frequently fails on realistic spatio-temporal queries. In our experiments, it serves both as a baseline for comparison and as the embedded SQL-generation subroutine within the agentic pipeline.

\subsection{Agentic Pipeline}
The agentic pipeline wraps the naive pipeline in a reasoning loop. It employs \textbf{Mistral Large}~\cite{mistralai2024large} orchestrated with \textbf{LangGraph}~\cite{langgraph2024} and follows a ReAct-style \emph{plan–act–observe} paradigm~\cite{yao2023react}. The key difference is that the agent can plan, decompose, and adapt queries by calling tools at will—including the naive SQLCoder pipeline.

\paragraph{Toolset.} The agent has access to six tools:
\begin{itemize}
    \item \textbf{\texttt{get\_database\_schema\_tool}} – retrieves schema metadata (tables, columns) and three sample rows from each table.
    \item \textbf{\texttt{generate\_sql\_query\_tool}} – wraps the naive SQLCoder baseline, producing SQL from a reformulated prompt.
    \item \textbf{\texttt{execute\_on\_database\_tool}} – runs SQL, saves results, and returns three sample rows for inspection.
    \item \textbf{\texttt{read\_file\_tool}} – reads large set of results from files.
    \item \textbf{\texttt{plot\_results\_tool}} – creates categorical or temporal plots.
    \item \textbf{\texttt{map\_results\_tool}} – generates point maps or density heatmaps.
\end{itemize}

\paragraph{Control flow.} A typical trajectory is:
\begin{enumerate}
    \item Parse the natural-language query and plan subtasks.
    \item Retrieve schema to ground generation.
    \item Call the naive SQLCoder tool to generate SQL.
    \item Execute SQL and inspect results.
    \item If results are incomplete, empty, or erroneous: retry with refinements (e.g., rephrase, broaden categories, substitute bounding boxes).
    \item Optionally visualize results (plot/map) and summarize them in natural language.
\end{enumerate}

Unlike the naive baseline, the agent may issue multiple queries, adapt predicates (e.g., defining “nightlife” as a set of categories), or inject external knowledge (e.g., borough boundaries, holiday dates). It can also decide that a visualization is more appropriate than a raw table.

\paragraph{Relationship to naive pipeline.} The naive pipeline is thus a \emph{subroutine}—an atomic SQL generation capability—inside the agentic system. The agent adds value by (i) rephrasing questions to align with schema, (ii) chaining multiple calls to decompose complex requests, (iii) recovering from execution errors, and (iv) delivering task-appropriate visualizations and summaries. This layered design makes the comparison fair: both systems rely on the same SQL model, isolating the effect of orchestration and tool use.

\paragraph{Design rationale.} By embedding the naive pipeline within an agent, we test the hypothesis that orchestration—rather than more powerful SQL generation alone—is the critical factor in handling realistic spatio-temporal queries.

\section{Evaluation Setup}
\textbf{Tasks and taxonomy.} We evaluate on a test set of \textbf{35} natural-language (NL) questions designed to reflect realistic analyst intents over spatio-temporal check-ins. Each question is annotated with one or more categories drawn from the following taxonomy (used throughout the paper and in Table~\ref{tab:per-question}):
\begin{itemize}
    \item \textbf{(B) Basic Filtering}: single-table selection with straightforward \texttt{WHERE} predicates (e.g., by user or category).
    \item \textbf{(A) Aggregation/Ranking}: grouping, counting, averaging, top-$k$, or max/min.
    \item \textbf{(T) Temporal Reasoning}: hour-of-day, weekday/weekend, holidays, or seasonal comparisons.
    \item \textbf{(M) Multi-step Reasoning}: decomposition into sub-queries or staged computation (e.g., "find top-$k$ then analyze them").
    \item \textbf{(S) Spatial/Geographic}: location-based predicates (places, bounding boxes, neighborhoods/boroughs).
    \item \textbf{(E) External Knowledge}: information not present in schema (e.g., holiday dates, neighborhood bounds, landmark coordinates).
    \item \textbf{(X) Multi-table/Dataset}: queries comparing or combining aligned datasets (e.g., NYC vs.\ Tokyo).
\end{itemize}
Coverage of the 35 queries is: \textbf{B} (6), \textbf{A} (26), \textbf{T} (19), \textbf{M} (7), \textbf{S} (7), \textbf{E} (6), \textbf{X} (5). The full questions and their category tags appear in Table~\ref{tab:per-question}.

\smallskip
\noindent\textbf{Systems compared.}
We compare two systems that use the \emph{same} SQL generator to isolate the effect of agentic orchestration:
\begin{enumerate}
    \item \textbf{Naive NL-to-SQL baseline}: \texttt{defog/llama-3-sqlcoder-8b} produces a single SQL statement directly from the user question plus the concrete schema (hard-coded in the prompt). No tool use, no retries, no rephrasing, no visualization planning.
    \item \textbf{Agentic pipeline}: A ReAct-style Mistral agent (LangGraph) that can call six tools at will: \texttt{get\_database\_schema\_tool}, \texttt{generate\_sql\_query\_tool} (which wraps the same SQLCoder model), \texttt{execute\_on\_database\_tool}, \texttt{read\_file\_tool}, \texttt{plot\_results\_tool}, and \texttt{map\_results\_tool}. The agent may reformulate questions using schema terms, decompose into sub-queries, retry after execution errors, and (optionally) produce plots/maps.
\end{enumerate}
Both systems run with temperature $0.0$. The underlying database is PostgreSQL \emph{without} geodesic extensions (e.g., no PostGIS); spatial filters use attribute/category predicates or axis-aligned bounding boxes when needed.

\smallskip
\noindent\textbf{Datasets and schema.}
Unless otherwise noted, queries operate on a single table \texttt{public.checkins} with fields \texttt{user\_id}, \texttt{place\_id}, \texttt{category\_name}, \texttt{latitude}, \texttt{longitude}, \texttt{checkin\_time}, etc.\ (see Dataset section). For multi-dataset queries (class \textbf{X}, Q31--Q35), we evaluate on two aligned tables (e.g., NYC and Tokyo) sharing the same schema.

\smallskip
\noindent\textbf{Protocol and success criteria.}
Each question is executed end-to-end by both systems. We mark an answer \emph{correct} if the produced SQL executes and the returned result set (or its clear textual/visual synthesis) \emph{semantically} answers the question:
\begin{itemize}
    \item \emph{Filtering queries} (B): rows match the intended predicate(s).
    \item \emph{Aggregation/ranking} (A): groupings and orderings are correct; ties and rounding are accepted if they do not change the stated conclusion.
    \item \emph{Temporal} (T): correct extraction/bucketing (e.g., hour-of-day) and correct holiday/season windows when applicable.
    \item \emph{Spatial} (S): correct place/category localization or bounding boxes when the schema lacks explicit regions.
    \item \emph{External knowledge} (E): correct incorporation of holiday dates or landmark/region bounds; we accept approximate bounds when they are reasonable and do not flip the conclusion.
    \item \emph{Multi-table} (X): correct per-dataset computation and cross-dataset comparison.
\end{itemize}
Visualizations are \emph{not} required for correctness unless explicitly requested; when produced, we log whether the chosen visual form is appropriate (used later in the qualitative analysis). All judgments were manually verified by the authors.

\section{Results}
\subsection{Quantitative Evaluation}
On the \textbf{35}-question benchmark, the agentic pipeline achieved \textbf{32/35} correctness (\textbf{91.4\%}), far surpassing the naive baseline at \textbf{10/35} (\textbf{28.6\%}). Per-question correctness is detailed in Table~\ref{tab:per-question}, with aggregate category-level success rates in Table~\ref{tab:category-success}.

The results highlight several clear patterns. The naive baseline performs reasonably only on \textbf{basic filtering} queries (50\% accuracy) and a subset of simple \textbf{aggregations} (26.9\%). However, it fails almost entirely on tasks requiring deeper reasoning: 0\% on spatial/geographic queries, 0\% on external knowledge queries, and 0\% on multi-dataset comparisons. In contrast, the agentic pipeline delivers consistently high accuracy across all categories, including \textbf{100\%} on multi-step reasoning (M) and nearly perfect performance on aggregation/ranking (96.2\%) and temporal reasoning (94.7\%). Substantial improvements are also seen in spatial/geographic tasks (0\% $\rightarrow$ 85.7\%), queries requiring external knowledge (0\% $\rightarrow$ 83.3\%), and queries requiring use of multiple tables (0\% $\rightarrow$80.0\%). 

\begin{table*}[t]
\centering
\caption{Per-question correctness (\cmark/\xmark) for naive vs.\ agentic pipelines on 35 spatio-temporal NL-to-SQL queries. Each query is annotated with its main categories: 
(B) Basic Filtering, 
(A) Aggregation/Ranking, 
(T) Temporal Reasoning, 
(M) Multi-step Reasoning, 
(S) Spatial/Geographic, 
(E) External Knowledge, 
(X) Multi-table/Dataset).}
\label{tab:per-question}
\small
\begin{tabular}{@{}r p{0.54\textwidth} p{0.14\textwidth} cc@{}}
\toprule
\# & User Question & Categories & Naive & Agentic \\
\midrule
1  & Show all check-ins at train stations. & B & \xmark & \cmark \\
2  & List check-ins made by user 123. & B & \cmark & \cmark \\
3  & Find all check-ins at restaurants in 2012. & B, T & \cmark & \cmark \\
4  & What are the top 10 most popular places? & A & \xmark & \cmark \\
5  & Which category had the most check-ins overall? & A & \cmark & \cmark \\
6  & How many check-ins happened each month in 2012? & A, T & \cmark & \cmark \\
7  & When during the day do people check in at coffee shops? & A, T & \xmark & \cmark \\
8  & Compare check-ins at 1pm vs 1am. & A, T & \cmark & \cmark \\
9  & Show weekday vs weekend check-in totals. & A, T & \xmark & \cmark \\
10 & Which category dominates late-night activity (10pm--4am)? & A, T & \xmark & \cmark \\
11 & For the top 5 categories, what is the average check-in time of day? & A, T, M & \xmark & \cmark \\
12 & For each of the top 3 categories, show the trend of monthly check-ins over 2012. & A, T, M & \cmark & \cmark \\
13 & Among the top 10 places, which has the largest share of late-night check-ins? & A, T, M & \xmark & \cmark \\
14 & For the top 5 users, show their distribution of activity across categories. & A, M & \xmark & \cmark \\
15 & Map the locations of all laundromats. & B, S & \xmark & \cmark \\
16 & Where are most gyms located? & A, S & \xmark & \cmark \\
17 & Show check-ins at JFK Airport. & S, E & \xmark & \xmark \\
18 & Compare check-ins in Midtown vs Downtown Manhattan. & M, S, E & \xmark & \cmark \\
19 & How do categories of check-ins differ between Brooklyn and Queens? & A, M, S, E & \xmark & \cmark \\
20 & Compare the number of morning check-ins versus evening check-ins at Central Park. & A, T, S & \xmark & \cmark \\
21 & Show all check-ins at Pizza Joints. & B & \xmark & \xmark \\
22 & Find all check-ins in February 2015. & B, T & \cmark & \cmark \\
23 & Show the busiest subway station. & A, S & \xmark & \cmark \\
24 & Show all check-ins at places with more than 1{,}000 visits. & A & \cmark & \cmark \\
25 & Were check-ins higher on New Year’s Eve than on an average day? & A, T, M, E & \xmark & \cmark \\
26 & Show check-ins during Thanksgiving Day. & T, E & \xmark & \cmark \\
27 & Compare summer vs winter check-in activity. & A, T, E & \xmark & \cmark \\
28 & Which places are busiest during lunch hours (11am--2pm)? & A, T & \xmark & \cmark \\
29 & Show trends in nightlife activity over time. & A, T & \cmark & \cmark \\
30 & How does check-in activity change across different times of day? & A, T & \cmark & \cmark \\
31 & Compare the most popular categories in NYC vs Tokyo. & A, M, X & \xmark & \cmark \\
32 & Which city has more late-night check-ins? & A, T, X & \xmark & \cmark \\
33 & Show the average check-ins per place in NYC vs Tokyo. & A, X & \xmark & \cmark \\
34 & Which city has more check-ins at train stations? & A, X & \xmark & \cmark \\
35 & Compare weekend activity patterns between NYC and Tokyo. & A, T, X & \xmark & \xmark \\
\midrule
\multicolumn{2}{r}{\textbf{Accuracy (correct/35)}} & & \textbf{10/35 (28.6\%)} & \textbf{32/35 (91.4\%)} \\
\bottomrule
\end{tabular}
\end{table*}

\begin{table}[t]
\centering
\caption{Success rates by query category. Queries can belong to multiple categories.}
\label{tab:category-success}
\small
\begin{tabular}{@{}lcc@{}}
\toprule
Category & Naive Accuracy & Agentic Accuracy \\
\midrule
Basic Filtering (B)       & 3/6 (50.0\%)  & 5/6 (83.3\%) \\
Aggregation/Ranking (A)   & 7/26 (26.9\%) & 25/26 (96.2\%) \\
Temporal Reasoning (T)    & 7/19 (36.8\%) & 18/19 (94.7\%) \\
Multi-step Reasoning (M)  & 1/7 (14.3\%)  & 7/7 (100\%) \\
Spatial/Geographic (S)    & 0/7 (0\%)     & 6/7 (85.7\%) \\
External Knowledge (E)    & 0/6 (0\%)     & 5/6 (83.3\%) \\
Multi-table/Dataset (X)   & 0/5 (0\%)     & 4/5 (80.0\%) \\
\midrule
Overall                   & 10/35 (28.6\%) & 32/35 (91.4\%) \\
\bottomrule
\end{tabular}
\end{table}

These findings underscore the advantages of schema awareness, decomposition, and tool use. The naive system, even though powered by the same SQL generation model, is limited to straightforward mappings of queries into SQL. The agent extends this baseline with the ability to reformulate, validate, and recover, which directly translates into dramatic gains on categories that mirror real-world analyst needs—spatio-temporal reasoning, multi-step decomposition, and cross-dataset synthesis.

\subsection{Qualitative Analysis}

\subsubsection{Rephrasing questions for SQL}
The agent reformulates vague natural-language requests into SQL-compatible prompts by adding concrete details absent from the user query. For example, for \emph{"Show trends in nightlife activity over time"}, the phrase "nightlife activity" does not appear in the schema and "over time" lacks a defined temporal resolution. Before generating SQL, the agent:
\begin{enumerate}
    \item Maps "nightlife activity" to explicit schema categories by filtering \texttt{Bar}, \texttt{Nightclub}, and \texttt{Music Venue}.
    \item Translates "over time" to a monthly aggregation via \newline\texttt{date\_trunc('month', checkin\_time)}.
\end{enumerate}
These additions bridge the semantic gap between human phrasing and database operations, yielding correct monthly counts and an interpretable line plot.

\subsubsection{Schema awareness and adaptation}
When asked to \emph{"Map the locations of all laundromats"}, the schema contained no "Laundromat" label. The agent explored \texttt{category\_name} values (e.g., substrings "laundry", "wash"), discovered the label \emph{"Laundry Service"}, and executed:
    \begin{verbatim}
    SELECT c.latitude, c.longitude
    FROM checkins c
    WHERE c.category_name = 'Laundry Service';
    \end{verbatim}
It then plotted 721 results as a heatmap, aligning user intent with schema representation and avoiding the failure modes of a naive pipeline.

\subsubsection{Decomposing tasks}
For \emph{Compare the number of morning check-ins versus evening check-ins at "Central Park"}, the agent decomposed the task:
\begin{enumerate}
    \item Defined Central Park via lat/lon bounds.
    \item Computed separate counts for morning (06:00--12:00) and evening (18:00--24:00).
    \item Synthesized the result: evening exceeded morning by 243 check-ins.
\end{enumerate}
Decomposition avoids brittle single-query formulations and supports interpretable summaries.

\subsubsection{Leveraging real-world knowledge}
For \emph{"How do categories of check-ins differ between Brooklyn and Queens?"}, the schema lacked a borough column. The agent injected geographic priors by providing approximate lat/lon bounds for each borough, guiding SQL generation to assign borough labels and aggregate by category, e.g.:
\begin{verbatim}
CASE 
  WHEN latitude BETWEEN 40.5707 AND 40.7395 
   AND longitude BETWEEN -74.0423 AND -73.8334 THEN 'Brooklyn'
  WHEN latitude BETWEEN 40.5091 AND 40.8007 
   AND longitude BETWEEN -73.9642 AND -73.7004 THEN 'Queens'
END AS borough
\end{verbatim}

\subsection{User Experience Enhancements}
\begin{figure}[h]
    \centering
    \includegraphics[width=0.92\linewidth]{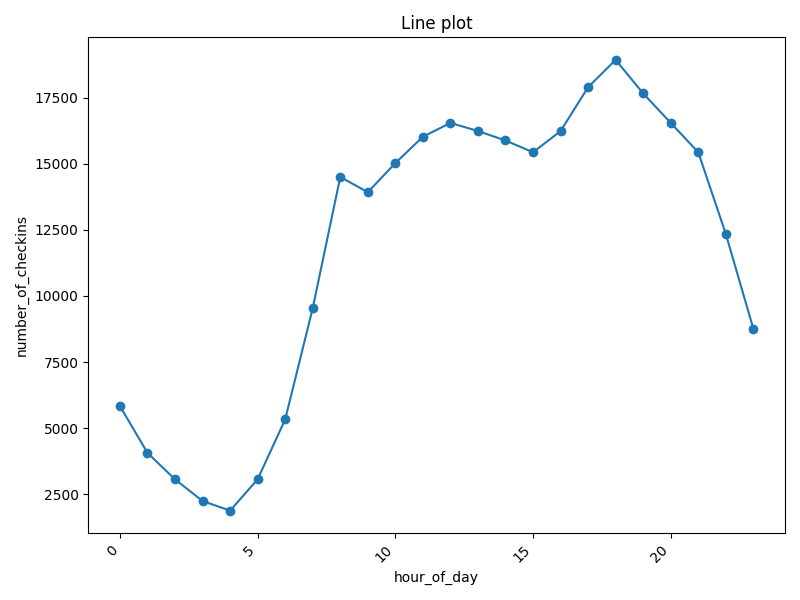}
    \caption{Hourly check-ins (line plot) generated by the agent for Q30: "How does check-in activity change across different times of day?". The agent produced this visualization without explicit request from the user.}
    \label{fig:hourly_line}
\end{figure}

\begin{figure}[h]
    \centering
    \includegraphics[width=0.85\linewidth]{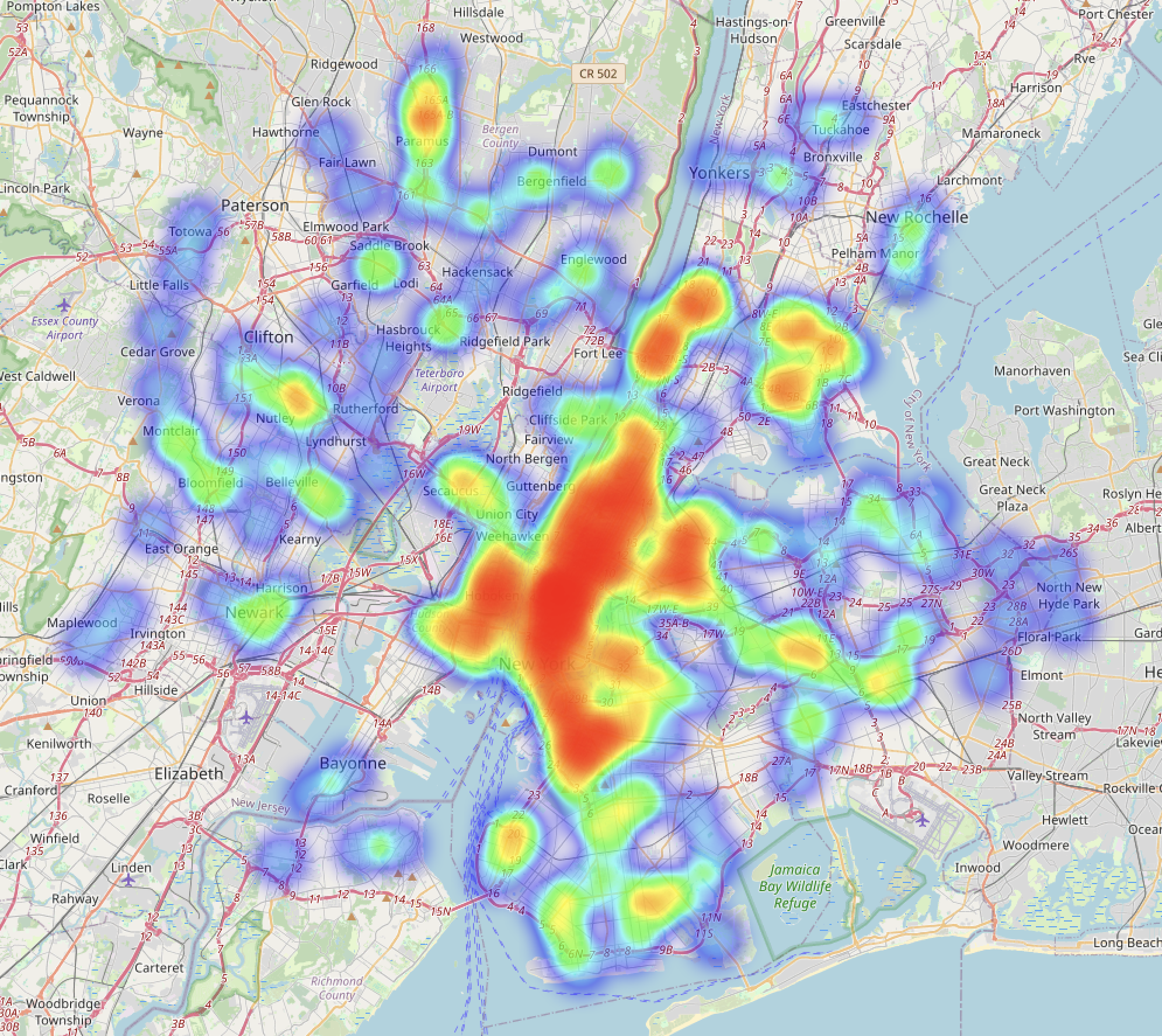}
    \caption{Agent-generated heatmap of gym locations for Q16: "Where are most gyms located?". The agent produced this visualization without explicit request from the user. Dynamic generation of a heatmap demonstrates the agent's ability to capture the intent of the user question.}
    \label{fig:map}
\end{figure}

Beyond accuracy, the agent improved usability by enriching outputs with both visual and textual aids. Naive text-to-SQL systems typically return only raw query results, leaving interpretation to the user. In contrast, the agent pipeline produced:

\begin{itemize}
    \item \textbf{Visualization}: automatic generation of maps, heatmaps, and plots (e.g., Figure~\ref{fig:hourly_line}, Figure~\ref{fig:map}), enabling direct perception of spatial and temporal trends without manual specification by the user.
    \item \textbf{Summarization and Insight}: structured natural-language explanations highlighting peaks, patterns, and comparisons. These reduced cognitive load by providing immediate interpretation of the results rather than forcing the user to infer them from tables or plots alone.
\end{itemize}

\noindent For example, when asked \emph{"How does check-in activity change across different times of day?"}, the agent not only generated a line plot but also produced a structured narrative organized into intuitive dayparts:

\begin{itemize}
    \item \textbf{Late Night (0--4 AM):} Minimal check-ins, bottoming out around 3--4 AM.  
    \item \textbf{Early Morning (5--7 AM):} Activity began to rise sharply around 6 AM, reflecting commuting and morning routines.  
    \item \textbf{Morning (8--11 AM):} A pronounced peak at 8 AM, sustained through late morning, likely tied to work and school arrivals.  
    \item \textbf{Midday (12--3 PM):} High activity through lunchtime, with a slight dip at 2 PM.  
    \item \textbf{Afternoon (4--6 PM):} Another peak at 6 PM, corresponding to end-of-work transitions and social activities.  
    \item \textbf{Evening (7--11 PM):} Activity remained elevated in early evening, peaking at 7 PM before tapering steadily toward midnight.  
\end{itemize}

\noindent This kind of structured textual summary illustrates how the agent goes beyond SQL execution: it transforms results into usable knowledge. A naive pipeline could, at best, output hourly counts in tabular form, leaving interpretation entirely to the user. By selecting appropriate visualizations and pairing them with interpretable textual narratives, the agent substantially reduces user effort in understanding, particularly for spatio-temporal data where trends unfold across time and space.

\section{Discussion}

\paragraph{Overall performance.}
The agentic pipeline substantially outperforms the single-pass text-to-SQL baseline (91.4\% vs.\ 28.6\%). These gains are consistent across temporal, spatial, and comparative queries, underscoring the importance of orchestration and adaptive reasoning even when both systems rely on the same underlying SQL generation model.

\paragraph{What drove the gains.}
Two design choices were decisive: (i) \emph{grounding and decomposition}—the agent reformulates underspecified prompts into schema-aligned predicates and decomposes multi-constraint questions into verifiable sub-queries; and (ii) \emph{usability}—the agent produces task-appropriate plots/maps and concise textual summaries, substantially reducing cognitive load compared to raw tabular outputs.

\paragraph{Error analysis: failure modes and remedies.}
Despite strong overall performance, the agent failed on three queries, illustrating distinct limitations:
\begin{itemize}
  \item \emph{Q17 (JFK).} The agent attempted to use radial distance filters around JFK airport, despite the database lacking geodesic functions (e.g., PostGIS). The generated query was not executable. \textbf{Remedies:} (a) detect unavailable functions and substitute axis-aligned bounding boxes, (b) maintain a dictionary of common landmark bounds, and (c) optionally leverage PostGIS when available.  
  \item \emph{Q21 ("Pizza Joints").} The agent overfit to the literal phrase "Pizza Joints", failing to generalize to schema labels such as "Pizza Place" or "Pizzeria." This was a rare error based on the rest of the results. \textbf{Remedies:} (a) perform two-step label discovery with fuzzy/substring search (e.g., \texttt{ILIKE '\%pizza\%'}), (b) maintain lightweight synonym lists, and (c) enforce exploratory label discovery when no direct match is found.  
  \item \emph{Q35 (Weekend NYC vs.\ Tokyo).} The agent hallucinated an unsupported approach to comparing weekend activity across two tables, producing incorrect logic despite access to the schema. This highlights a limitation of planning reliability when reasoning across datasets. \textbf{Remedies:} (a) constrain multi-table comparisons to validated join patterns, (b) check generated queries for unsupported constructs before execution, and (c) add explicit retry strategies for cross-dataset aggregation failures.  
\end{itemize}

\paragraph{Trade-offs.}
While the agent improves correctness and usability, orchestration introduces additional computation. In our evaluation, the naive pipeline made exactly one call to the SQL generation model per query, whereas the agent averaged \textbf{1.51 calls}. Both used the same SQL generator (\texttt{defog/llama-3-sqlcoder-8b}) running on an NVIDIA L4 GPU. The agent itself ran on the Mistral API, which contributed negligible latency; the dominant latency factor was SQL generation. Thus, the cost of improved accuracy is modest: approximately 50\% more SQL generation calls per query. Latency and cost can be mitigated by caching schema metadata, discovered categories, and common query templates, while selectively bypassing visualization when it adds little value.

\paragraph{Threats to validity.}
Results are based on one dataset and one baseline. Geographic priors (e.g., bounding boxes) were manually specified, and correctness was judged by semantic sufficiency rather than strict SQL equivalence. Broader evaluations across datasets, schemas, and baseline models would strengthen the generality of our findings.

\section{Conclusion and Future Work}
This paper introduced and evaluated an agentic LLM pipeline for spatio-temporal text-to-SQL, applied to the NYC and Tokyo check-in dataset. By embedding a naive SQL generation model within a ReAct-style agent equipped with schema inspection, execution, visualization, and error-recovery tools, we isolated the impact of orchestration over raw model strength. Across a benchmark of 35 realistic queries, the agentic pipeline achieved \textbf{91.4\%} correctness, far exceeding the \textbf{28.6\%} of a naive baseline that used the same SQL generator in isolation. These results demonstrate that orchestration—rather than more powerful text-to-SQL models alone—enables robust performance on complex spatio-temporal queries.

Our findings highlight three central takeaways. First, \emph{schema-aware reformulation and decomposition} were essential to bridging gaps between natural phrasing and structured queries, especially for temporal and multi-step reasoning tasks. Second, the agent’s ability to \emph{adapt and recover from errors} yielded major gains in categories where the baseline failed completely, including spatial filtering, external knowledge, and cross-dataset synthesis. Third, usability was substantially improved through \emph{automatic visualizations and structured textual summaries}, which transformed raw results into interpretable knowledge and reduced cognitive effort for non-expert users.

At the same time, failure settings emerged: tendency to use unsupported geodesic functions (Q17), over-literal label matching without synonym expansion (Q21), and planning instability with complex cross-dataset synthesis (Q35). These illustrate the limits of current orchestration and motivate safeguards such as synonym discovery, query linting, and constrained multi-table joins.

\subsection{Practical Guidance}
From our evaluation, we distill four key principles for designing agentic NL-to-SQL systems:

\begin{enumerate}
    \item \textbf{Ground intent in schema and context.} Translate vague or colloquial phrasing into schema-aligned predicates (e.g., map "nightlife" to \texttt{Bar}, \texttt{Nightclub}, \texttt{Music Venue}), and default to bounding boxes when spatial functions are unavailable. Maintain small dictionary and synonym lists to reconcile user vocabulary with schema values.  

    \item \textbf{Use errors and discovery as feedback.} Treat empty results or execution failures as signals to retry with refined strategies, such as exploratory label discovery (\texttt{DISTINCT category\_name} with fuzzy matching) before filtering.  

    \item \textbf{Decompose and synthesize.} Break down multi-step requests into sub-queries, execute them independently, and then combine results. Pair outputs with sensible default visualizations (line for temporal, bar for categorical, heatmap for spatial) and concise textual summaries to aid sensemaking.  

    \item \textbf{Balance efficiency with reliability.} Validate queries before execution to catch unsupported constructs, and cache schema metadata, discovered labels, and templates. Although orchestration increased SQL generation calls modestly (1.51 on average vs.\ 1.0 for the naive baseline), caching and pre-validation mitigate latency and cost. The ReAct-style \newline plan–act–observe loop remains central for refining underspecified or error-prone queries.  
\end{enumerate}

\subsection{Future Work}
Several directions remain open for advancing agentic NL-to-SQL pipelines:
\begin{enumerate}
    \item \textbf{Scale and generalization.} Extend evaluation beyond NYC to Tokyo and other heterogeneous spatio-temporal datasets, testing generalization to unseen schemas.  
    \item \textbf{Ablations and tool attribution.} Quantify the marginal contribution of each tool (schema inspection, retries, visualizations) and identify minimal tool sets for robust performance.  
    \item \textbf{Interactive refinement.} Incorporate conversational memory and user feedback to support iterative correction, preference learning, and cooperative query design.  
    \item \textbf{Efficiency.} Reduce overhead by learning policies for when to call tools, distilling orchestration strategies, and caching common queries or visualizations.  
    \item \textbf{Reliability.} Improve planning robustness in cross-dataset synthesis and implement systematic guardrails against unsupported constructs or hallucinated joins.  
\end{enumerate}

Taken together, these directions move toward geospatial assistants that are not only more accurate but also more efficient, reliable, and user-centered. Our results provide evidence that agentic orchestration represents a practical and impactful path forward for natural-language interfaces to spatio-temporal databases.

\bibliographystyle{ACM-Reference-Format}
\bibliography{references}

\end{document}